\documentclass{article} 
\usepackage{iclr2027_conference,times}

\usepackage{amsmath,amsfonts,bm}

\def\eqref#1{equation~\ref{#1}}

\def\1{\bm{1}}

\DeclareMathAlphabet{\mathsfit}{\encodingdefault}{\sfdefault}{m}{sl}
\SetMathAlphabet{\mathsfit}{bold}{\encodingdefault}{\sfdefault}{bx}{n}

\usepackage{hyperref}
\usepackage{url}
\usepackage{graphicx}
\usepackage{booktabs}
\usepackage{multirow}
\usepackage{amsmath}
\usepackage{amssymb}
\usepackage{xcolor}
\usepackage{wrapfig}

\newcommand{\student}[1]{S_{#1}}          
\newcommand{\teacher}[1]{T_{#1}}          

\title{GRAFT: Growing Agglomerative Foundation Models via Continual Teacher Distillation}

\author{
    Zhenghao Zhao\textsuperscript{1}\thanks{Work done during an internship at Amazon.} \qquad Chi Zhang\textsuperscript{2} \qquad Qingshuang Chen\textsuperscript{2} \qquad Yelin Kim\textsuperscript{2} \\
    \textsuperscript{1}University of Illinois Chicago \quad
    \textsuperscript{2}Amazon\\
    {\tt\small zzhao72@uic.edu}\\
    {\tt\small \{zhanbchi, cicichen, kimyelin\}@amazon.com}
}

\iclrfinalcopy 
\begin{document}

\maketitle
\lhead{Preprint.}

\begin{abstract}
Vision foundation models such as DINOv2, SigLIP2, and MASt3R develop complementary capabilities from different pretraining objectives, yet their knowledge remains distributed across separate, specialized models.
Multi-teacher knowledge distillation offers a path toward consolidating these capabilities into a single agglomerative backbone, but existing approaches assume a fixed set of teachers, and incorporating a new teacher requires repeating expensive joint distillation over the entire teacher set.
We introduce \textbf{GRAFT}, a continual multi-teacher distillation framework
that enables a unified backbone to progressively acquire capabilities from an open-ended sequence of foundation models. When a new teacher arrives, GRAFT treats the previously distilled model as a teacher for preserving learned capabilities, while the current student jointly learns from both the previous model and the incoming teacher.
Furthermore, to reconcile the incompatible representation geometries of heterogeneous teachers, we introduce Teacher Specific Readout Tokens, which grant each teacher an independent read-out of the shared encoder, together with Geometry Agnostic Relational Loss that aligns a vision--language teacher by matching image–text similarity structures rather than raw feature values. 
We provide GRAFT model, which is a single, continually extensible backbone that unifies five domains, including image understanding, 2D dense prediction, 3D human pose estimation, 3D vision, and vision--language, delivering strong performance across all of them while acquiring each new capability at the cost of a single distillation rather than a full re-distillation.
\end{abstract}

\section{Introduction}
\label{sec:intro}
Progress in visual representation learning has produced an increasingly diverse ecosystem of foundation models~\citep{oquab2024dinov2,radford2021clip,tschannen2025siglip2,kirillov2023sam,leroy2024mast3r,baradel2024multihmr}. Rather than converging toward a single generalist representation, these models develop complementary capabilities from different pretraining objectives and data sources.
DINOv2 provides strong generic visual features, SigLIP2 aligns visual representations with language, MASt3R captures 3D geometry, and specialized models such as SAM target segmentation.
Deploying a separate encoder for every capability, however, quickly becomes impractical when memory, computation, or inference latency is constrained.
Agglomerative models such as AM-RADIO \citep{ranzinger2024radio}, DUNE \citep{sariyildiz2025dune}, SAK \citep{lu2025sak} and Theia \citep{shang2024theia} 
demonstrate that representations distilled from heterogeneous foundation models can coexist within a single encoder, typically through teacher specific projection heads and mechanisms for balancing multiple distillation objectives~\citep{ranzinger2024phis}.

However, what existing agglomerative models shares is that the teacher set is fixed.
Once training is complete, incorporating a newly available foundation model generally requires performing joint distillation again over the enlarged teacher set.
This assumption conflicts with how foundation models evolve in practice.
A stronger teacher is released, or a task the system did not originally require becomes necessary, and the desirable capability is one that appears after the student was trained. Adding it to an existing agglomerative student is not incremental under current methods. The student must be re-distilled from scratch against the enlarged teacher set, a joint MTKD run that keeps every earlier teacher live.
Reaching a target set of capabilities this way costs a full distillation run for every capability added, rather than one teacher's worth of work per step.

\begin{wrapfigure}{r}{0.45\textwidth}
    \centering
    \includegraphics[trim=0 10 0 10,clip,width=1\linewidth]{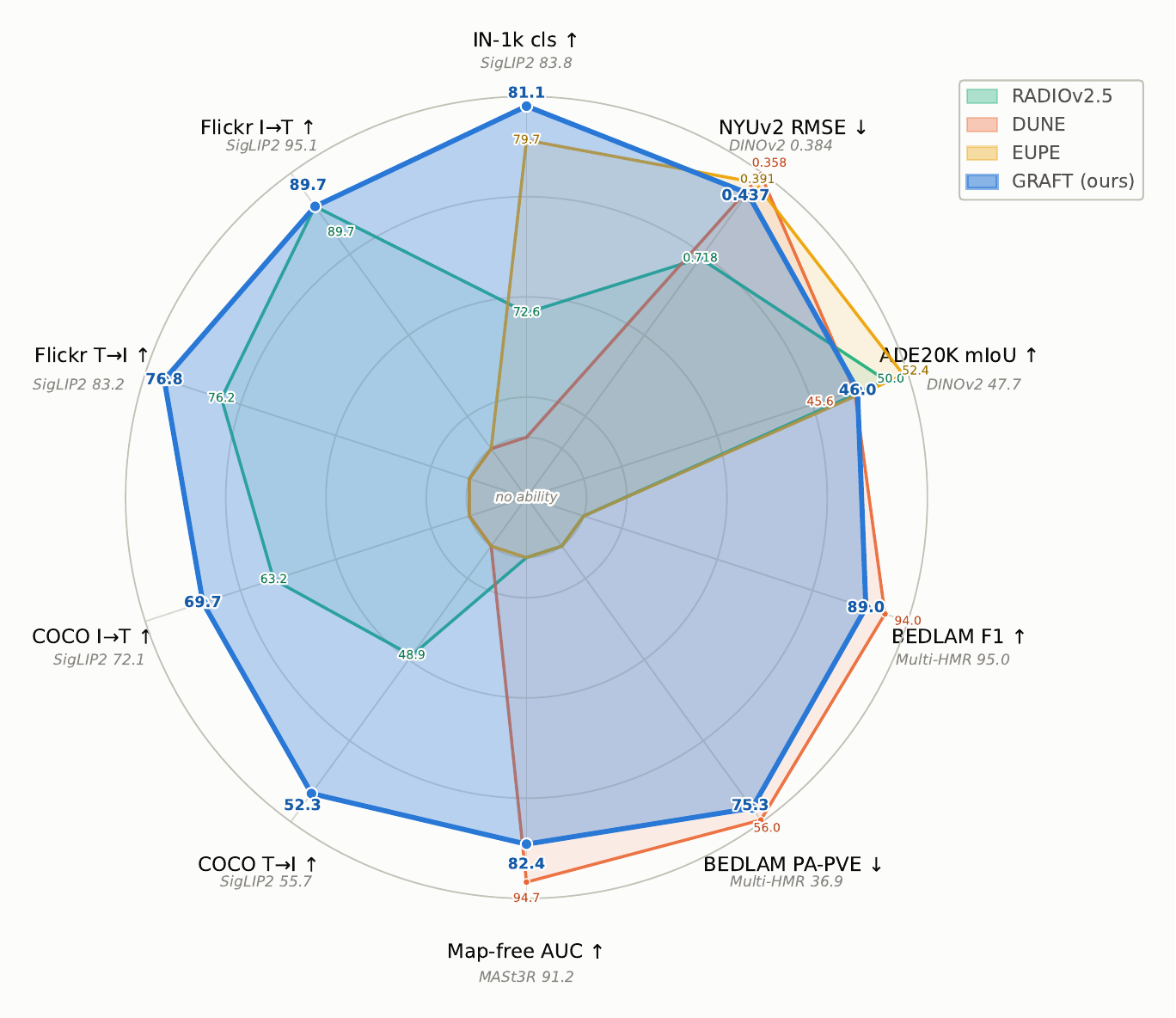}
    \vspace{-0.8cm}
    \caption{
    Compared to existing agglomerative models, GRAFT hold a comprehensive capability on various tasks with strong performance on each task.
    }
    \label{fig:cmtkd}
    \vspace{-10pt}
\end{wrapfigure}
\begin{figure}[t]
\begin{center}
\includegraphics[trim=5 0 5 20,clip,width=1\linewidth]{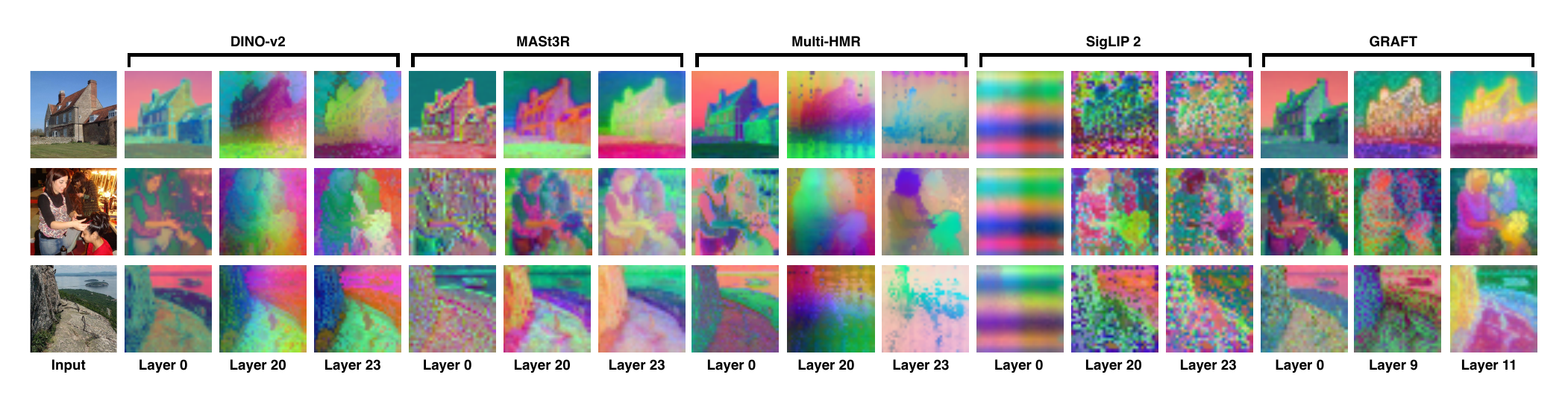}
\vspace{-1cm}
\end{center}
\caption{
\textbf{Per-patch feature geometry across depth.} Each cell visualizes the top-3 PCA components of patch tokens (mapped to RGB, fit per cell) at a shallow, deep, and final block for four ViT-L teachers and GRAFT (ViT-B) on three ADE20K images. The teachers diverge from the shallowest blocks: DINOv2 resolves object structure, MASt3R shows geometric texture, SigLIP2 carries only positional patterns, and Multi-HMR progressively organizes around people. GRAFT unifies these in a single trunk---its patch tokens retain sharp object/part structure from its dense teachers (DINOv2, MASt3R, Multi-HMR), while SigLIP2's vision--language capability is grafted relationally onto a dedicated summary token (\S\ref{sec:method:objectives}) and thus does not appear in this patch view.
}\vspace{-0.6cm}
\label{fig:pca}
\end{figure}

We study this setting as \textbf{continual multi-teacher distillation}.
Given a student that has already absorbed several foundation models, and an incoming teacher, our goal is to incorporate the new teacher without accessing any of the original teachers and without restarting distillation from scratch.
Meeting this goal 
raises two challenges. The first is \emph{catastrophic forgetting}: because the earlier teachers are discarded, any drift of the shared trunk toward the incoming teacher silently erodes the tasks the student already supports, and no original teacher remains to repair the damage. The second is \emph{teacher heterogeneity}: foundation models trained with different objectives can organize their representation spaces in substantially different ways. 
Forcing all of them to match a common representation through the same pointwise feature objective can introduce conflicting supervision. As illustrated in Figure~\ref{fig:pca}, this heterogeneity can emerge early in the network rather than only at the final representation layer.

We introduce \textbf{GRAFT}, a continual multi-teacher distillation framework that addresses forgetting through a key observation: 
the previous student already encodes the capabilities acquired from earlier teachers and can therefore serve as a compact surrogate for them during subsequent distillation.
At each round, GRAFT initializes the student from the previously distilled model, freezes a copy of the distilled model as an {anchor}, and jointly distills from this previous model and the incoming teacher.
The anchor model provides an {anchor loss} that discourages drift from previously acquired representations, while the incoming teacher provides supervision for the new capability. Each graft step therefore requires only two frozen reference models, the previous student and the incoming teacher, regardless of how many capabilities have already been accumulated.
Historical knowledge is carried forward inside the student itself rather than through an ever growing buffer of past teachers.

GRAFT further addresses teacher heterogeneity at both the architectural and objective levels. Architecturally, we introduce {Teacher Specific Readout Tokens}, which provide each teacher with an independent global readout from the shared encoder. Each readout is paired with its own projector and normalization parameters, allowing heterogeneous teachers to extract different global summaries from a common trunk without forcing their representations into a single shared coordinate system. At the objective level, 
for vision-language teachers, we introduce a {Geometry Agnostic Relational Loss} that 
matches the image-text similarity structures rather than match teacher and student feature values directly. This allows the student to preserve the teacher's image text alignment while retaining a feature space compatible with previously acquired capabilities.

Together, we turn capability consolidation from a one time operation into a continual process. GRAFT progressively grows a single shared backbone by absorbing one foundation model at a time, requiring only a single incremental distillation stage for each incoming teacher. We evaluate the resulting model across five capability families spanning image understanding, 2D dense prediction, 3D human pose estimation, 3D vision, and vision language understanding. Compared with existing agglomerative models, GRAFT supports a comprehensive range of heterogeneous capabilities while maintaining strong performance across individual tasks, with the additional ability to continually incorporate new capabilities without full joint re-distillation.

Our contributions are summarized as follows:
\begin{itemize}
  \item We formalize {continual multi-teacher distillation} for vision
    foundation models, a setting in which heterogeneous teachers arrive over time
    with no access to prior teachers.
  \item We introduce \textbf{GRAFT}, a continual distillation framework that addresses both forgetting and teacher heterogeneity. GRAFT uses the previously distilled student as a frozen retention teacher to preserve accumulated capabilities while learning from each incoming teacher. To accommodate heterogeneous representations within a shared encoder, GRAFT further introduces {Teacher Specific Readout Tokens} for independent teacher specific readouts and a {Geometry Agnostic Relational Loss} that transfers vision-language knowledge by matching image text similarity structures rather than raw feature values.
  \item We demonstrate a distilled model spanning five capability families, including image understanding, 2D dense prediction, 3D human pose estimation, 3D vision, and vision language understanding, and show that GRAFT can acquire new capabilities while retaining previously distilled ones without full joint re-distillation over the accumulated teacher set.
\end{itemize}

\section{Related Work}
\label{sec:related}

\subsection{Vision foundation models}
Vision foundation models acquire markedly different representational biases from their pretraining objectives and data. Self-supervised visual pretraining produces broadly transferable representations with strong spatial structure, supporting tasks such as semantic segmentation and monocular depth estimation~\citep{oquab2024dinov2,simeoni2025dinov3}. Image--text contrastive pretraining instead emphasizes semantic alignment across modalities, enabling zero-shot recognition and cross-modal retrieval~\citep{radford2021clip,zhai2023siglip,tschannen2025siglip2,fang2024dfn}. Other models specialize more explicitly: SAM learns promptable segmentation~\citep{kirillov2023sam}, DUSt3R and MASt3R target multi-view geometry and 3D reconstruction~\citep{dust3r2024,leroy2024mast3r}, and MultiHMR targets multi-person human mesh recovery~\citep{baradel2024multihmr}. Despite their broad individual transferability, these models exhibit complementary strengths across task families rather than a single representation uniformly dominating across all capabilities. This complementarity motivates methods that consolidate multiple foundation models into a shared encoder.

\subsection{Agglomerative models and multi-teacher distillation}

Multi-teacher knowledge distillation provides a natural route for consolidating complementary foundation models into a shared backbone. AM-RADIO~\citep{ranzinger2024radio} introduced the agglomerative model paradigm, using teacher-specific heads to distill global and spatial representations from multiple teachers into a common ViT. Subsequent methods~\citep{sariyildiz2024unic,sariyildiz2025dune,lu2025sak,ranzinger2024phis,shang2024theia,heinrich2025radiov25,chaybouti2026siglino} expand the teacher set and improve how heterogeneous representations are reconciled. PHI-S~\citep{ranzinger2024phis} addresses mismatched teacher feature statistics through isotropic standardization. DUNE~\citep{sariyildiz2025dune} distills 2D, 3D, and human-perception teachers using teacher-specific transformer projectors.
SAK~\citep{lu2025sak} preserves teacher-specific biases through separate adaptation paths.

Multi-teacher distillation has also been explored beyond general-purpose agglomerative encoders. Related work integrates multiple visual teachers into vision--language models through lightweight adaptation modules~\citep{cao2025movekd,wang2025hawaii}, distills foundation models using compact proxy datasets for greater data efficiency~\citep{zhang2025proteus,lee2025customkd,frank2025goodkd}, studies heterogeneous and task-agnostic transfer objectives~\citep{li2025fusebeforetransfer,chen2025adm,busbridge2025distillscaling,formont2025taskagnostic}, and consolidates domain-specific teachers for medical imaging, 3D perception, and robotics~\citep{grashei2025pathryoshka,zeng2025vfss,li2025chorus,wang2026ver}. These approaches broaden the scope of multi-teacher distillation, but existing agglomerative formulations predominantly optimize over a teacher collection available during the corresponding distillation procedure. 

Despite these advances, existing methods are designed for a fixed teacher set. Adding a teacher after training therefore requires revisiting previous teachers or performing another joint consolidation stage. GRAFT instead treats the agglomerative backbone as continually extensible.

\subsection{Continual learning}
Preventing degradation of earlier capabilities is the central problem of
continual learning. Regularization methods penalize movement of parameters
important to previous tasks \citep{kirkpatrick2017ewc}; distillation-based
methods constrain the model's outputs toward those of its former self
\citep{li2018lwf}; replay methods retain or synthesize data from earlier tasks
\citep{rebuffi2017icarl}. 
SE2D~\citep{michel2026continual} study teachers trained on different domains and identify both the transfer of knowledge absent from the student's current data and its subsequent forgetting as central challenges. This establishes that sequential teacher distillation can be performed without retaining the full history of teacher models. 
While SE2D preserves predictive knowledge by matching logits from sequential teachers trained on different domains, GRAFT 
continually grows an agglomerative model whose teachers contribute fundamentally different representation-level capabilities, which span image understanding, dense prediction, 3D human pose, 3D geometry, and vision--language understanding, and therefore cannot generally be represented through a common logit space.
\section{Method}
\label{sec:method}

\subsection{Problem definition: Continual multi-teacher distillation}
\label{sec:method:problem}

We formalize \textbf{continual multi-teacher knowledge distillation} (continual MTKD). Heterogeneous vision teachers $\teacher{1}, \teacher{2}, \ldots$ arrive one at a time, starting from an initial student $\student{1}$, an arbitrary entry point such as a randomly initialized network or an off-the-shelf pretrained checkpoint. When teacher $\teacher{k}$ arrives, the learner must extend the previous student $\student{k-1}$ into a student $\student{k}$ that supports every task seen so far, $1, \ldots, k$, at the cost of distilling that single teacher.

At round $k$, the learner is given exactly three inputs:
\begin{itemize}
  \item the prior student $\student{k-1}$, a shared-trunk encoder that already supports tasks $1, \ldots, k-1$;
  \item an image corpus $\mathcal{D}_k$;
  \item the incoming teacher $\teacher{k}$, accessible only as a frozen feature extractor.
\end{itemize}
The defining constraint is that the past teachers $\teacher{1}, \ldots, \teacher{k-1}$ are unavailable.
Each student is evaluated by acquisition and retention. Acquisition measures how  effectively the new task $k$ is assimilated, and retention measures how robustly the earlier tasks $1, \ldots, k-1$ are preserved.
In the following sections, we first articulate the inherent challenges of continual MTKD, and subsequently present our proposed framework, GRAFT, to overcome them.

\begin{figure}[t]
\begin{center}
\includegraphics[width=1\linewidth]{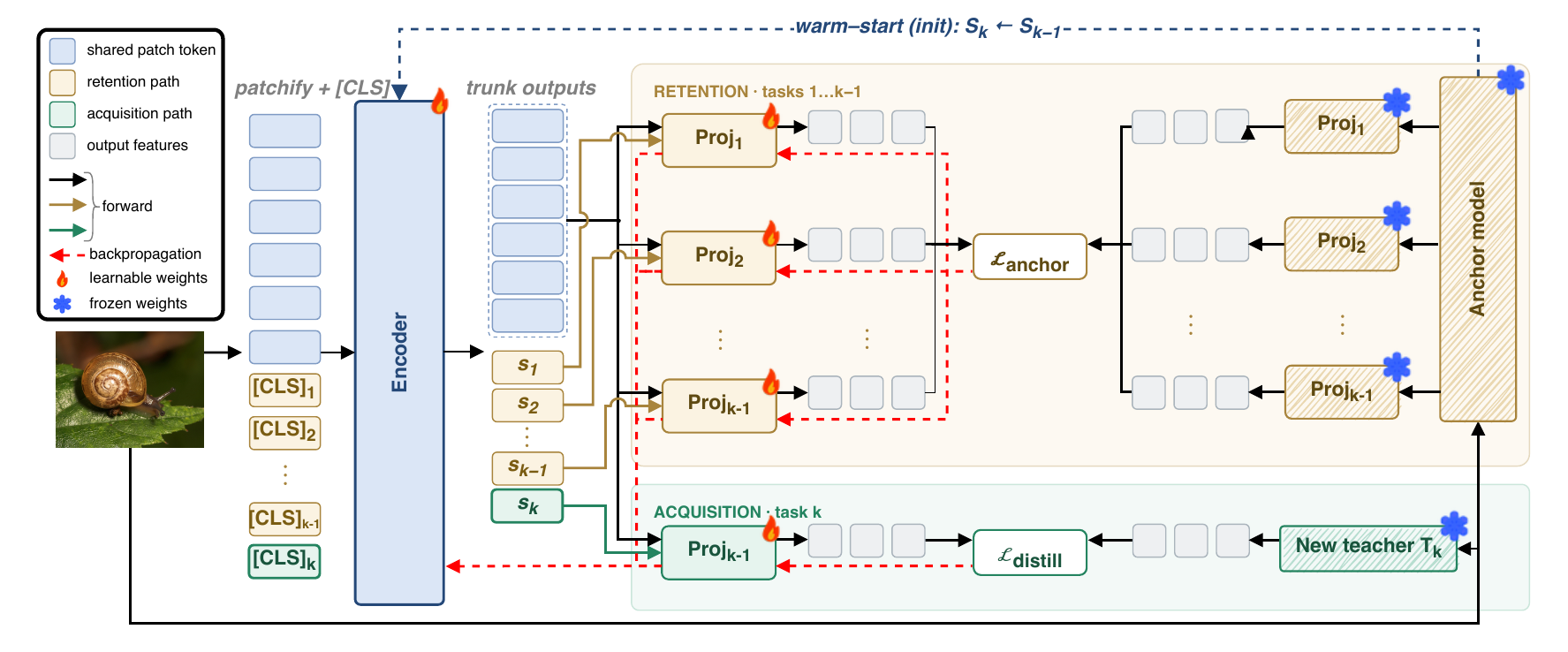}
\vspace{-1.2cm}
\end{center}
\caption{
\textbf{Overview of GRAFT at round $k$.}
The active student $\student{k}$ is warm-started from $\student{k-1}$.
A learnable Teacher Specific Readout Token (TSRT) is assigned to each task and jointly encoded with the shared patch tokens, producing task specific summary representations that are mapped by task-specific projectors.
Training follows two concurrent paths. In the {retention path}, outputs for previously acquired tasks $1,\ldots,k-1$ are constrained by $\mathcal{L}_{\mathrm{anchor}}$ to remain consistent with those of the frozen anchor model $\student{k-1}$, which preserves previously acquired task functions. In the {acquisition path}, the new task readout is optimized through $\mathcal{L}_{\mathrm{distill}}$ to acquire knowledge from the incoming teacher $\teacher{k}$. Only the active student $\student{k}$ is updated, while both $\student{k-1}$ and $\teacher{k}$ remain frozen.
}
\vspace{-0.4cm}
\label{fig:pipeline}
\end{figure}

\subsection{The GRAFT pipeline}
\label{sec:method:pipeline}

Continual multi-teacher knowledge distillation (MTKD) inherently suffers from catastrophic forgetting. As the model distills knowledge from a new teacher, the absence of supervision from discarded previous teachers causes the representation to drift toward the current task. 
We address this problem with {GRAFT}, a continual MTKD framework that uses the previously distilled student as a compact functional proxy for all past teachers. Because the previous student consolidates the functions acquired from earlier teachers, it can provide retention supervision.

Consider continual distillation over a sequence of teachers ${\teacher{1},\ldots,\teacher{k}}$. At round $k$, GRAFT maintains only the incoming teacher $\teacher{k}$ and the anchor model $\student{k-1}$, which is a frozen student obtained from round $k-1$. The active student $\student{k}$ is warm started from  $\student{k-1}$.
$\student{k}$ is then optimized to simultaneously acquire the new teacher while preserving the learned task functions.
Given an input image $\mathbf{x}$, we denote the student output associated with task $j$ at round $k$ as
\begin{equation}
\mathbf{s}_j^{(k)}=\mathcal{H}_j^{(k)}\left(\mathcal{E}^{(k)}(\mathbf{x})\right),
\label{eq:task-readout}
\end{equation}
where $\mathcal{E}$ denotes the shared visual backbone and $\mathcal{H}_j$ denotes the task specific projector associated with task $j$. 
This formulation allows different tasks to employ distinct readouts and distillation objectives while sharing the same visual backbone.

Learning is decomposed into two complementary objectives. The {acquisition objective} transfers knowledge from the incoming teacher $\teacher{k}$ into the newly instantiated task projector:
\begin{equation}
\mathcal{L}_{\mathrm{distill}}^{(k)}=d\left(\mathbf{s}_k^{(k)},\teacher{k}(\mathbf{x})\right),
\label{eq:acquisition-loss}
\end{equation}
where $d(\cdot,\cdot)$ denotes the distillation objective.
In parallel, the {retention objective} preserves the functions learned from all previous teachers. For every previous task $j < k$, the current student is constrained to reproduce the response of the corresponding readout in the frozen anchor model:
\begin{equation}
\mathcal{L}_{\mathrm{anchor}}^{(k)}=\sum_{j=1}^{k-1}\lambda_jd\left(\mathbf{s}_j^{(k)},\mathcal{H}_j^{(k-1)}\left(\mathcal{E}^{(k-1)}(\mathbf{x})\right)\right),
\label{eq:anchor-loss}
\end{equation}
where $\mathcal{E}^{(k-1)}$ and $\mathcal{H}_j^{(k-1)}$ denote the backbone and projector for task $j$ of anchor model $\student{k-1}$, and all parameters of $\student{k-1}$ are frozen. 
This objective regularizes the student in function space by maintaining the outputs of all previously instantiated task readouts.
The complete objective at continual round $k$ is
\begin{equation}
\mathcal{L}^{(k)}=\underbrace{d\left(\mathbf{s}_k^{(k)},\teacher{k}(\mathbf{x})\right)}_{\text{acquisition}}+\underbrace{\sum_{j=1}^{k-1}\lambda_jd\left(\mathbf{s}_j^{(k)},\mathcal{H}_j^{(k-1)}\left(\mathcal{E}^{(k-1)}(\mathbf{x})\right)\right)}_{\text{retention}},
\label{eq:graft-objective}
\end{equation}
where $\lambda$ controls the strength of historical preservation. The first term expands the student's capability toward the incoming teacher, while the second explicitly constrains previously learned task functions from drifting during this update. GRAFT therefore casts continual MTKD as an additive learning objective in which acquisition of new knowledge is regularized by preservation of previously consolidated functions.

A key property of this formulation is that historical supervision is compressed into a single anchor model. At round $k$, GRAFT requires the incoming teacher $\teacher{k}$, and the frozen historical student $\student{k-1}$, rather than the complete collection of past teachers ${\teacher{1},\ldots,\teacher{k-1}}$. Consequently, the number of large backbone models required for retention is independent of the number of previous teachers. 
The following subsection addresses a complementary challenge: heterogeneous teachers can impose conflicting representational requirements on the shared backbone. We therefore introduce task specific readout tokens and geometry agnostic objectives to isolate these heterogeneous interfaces.

\subsection{Addressing representation heterogeneity}
\label{sec:method:objectives}

\paragraph{Teacher Specific Readout Tokens.}
A straightforward multi teacher architecture attaches $k$ task specific projection heads to a single global student representation. 
Although the projection heads are independent, all tasks must read from the same global summary vector $\mathbf{z}$. This creates a representational bottleneck when the corresponding teachers encode substantially different semantics or organize their representation spaces differently.

To disentangle these heterogeneous readouts while retaining a shared visual representation, we introduce {Teacher Specific Readout Tokens} (TSRT). Instead of using a single global summary token, we associate each task $j$ with an independent learnable readout token $\mathbf{c}_j$. For an input image $\mathbf{x}$, all $k$ readout tokens are prepended jointly to the shared visual patch tokens $v(\mathbf{x})$:
\begin{equation}
    \mathbf{Z}_{\mathrm{in}}=\big[\mathbf{c}_1,\ldots,\mathbf{c}_k,v(\mathbf{x})\big].
\end{equation}
Then $\mathbf{Z}_{\mathrm{in}}$ is processed by the shared student backbone in the forward pass:
\begin{equation}
    \big[\tilde{\mathbf{c}}_1,\ldots,\tilde{\mathbf{c}}_k,\mathbf{Z}_{\mathrm{patch}}\big]=\mathcal{E}^{(k)}\big(\mathbf{Z}_{\mathrm{in}}\big),
\end{equation}
where $\tilde{\mathbf{c}}_j$ denotes the contextualized readout token associated with task $j$, and $\mathbf{Z}_{\mathrm{patch}}$ denotes the shared contextualized patch representation.
Each task then reads from the shared representation through its corresponding projector. Specifically, the projector receives its dedicated readout token together with the common patch representation:
\begin{equation}
    \mathbf{s}_j^{(k)}=\mathcal{H}_j^{(k)}\big(\tilde{\mathbf{c}}_j\oplus\mathbf{Z}_{\mathrm{patch}}\big).
\end{equation}
This architecture therefore shares the visual computation while factorizing the global readout across tasks. 
A single global readout forces these heterogeneous objectives to compete through the same summary representation, even when the underlying visual features useful to them can be shared. TSRT instead separates the shared visual representation from the task dependent readout, allowing each task to summarize the common patch features according to its own distillation objective.

\paragraph{Geometry Agnostic Relational Loss.}
Vision-language teachers introduce a distinct challenge in heterogeneous multi teacher distillation because their representation geometry can differ substantially from that of dense or geometric vision teachers.
As illustrated by the PCA visualization in Figure~\ref{fig:pca}, the representations produced by these encoders diverge from the shallowest layers onward.
A conventional feature distillation objective directly aligns the student and teacher image embeddings.
Such an objective implicitly assumes that direct correspondence between the student and teacher feature spaces is desirable. 
However, enforcing this correspondence can 
perturb features required by previously distilled tasks.

We therefore avoid directly matching the image embeddings of the vision language teacher. Instead, we exploit the property for which such a teacher is particularly valuable: its learned alignment between images and language. 
Give student and supervising feature outputs $\mathbf{s}_j^{}$ and $\hat{\mathbf{s}}_j^{}$
we require the student and teacher to induce similar relationships between image and text features:
\begin{equation}
\operatorname{sim}\left(\mathbf{s}_j^{},\mathbf{t}\right)\approx\operatorname{sim}\left(\hat{\mathbf{s}}_j^{},\mathbf{t}\right),
\label{eq:relational-intuition}
\end{equation}
where $\mathbf{t}$ denotes the text features produced by the frozen text encoder. We convert these image text similarities into probability distributions:
\begin{equation}
    p_{j}^{t} =\operatorname{softmax}
    \left(
        \frac{\hat{\mathbf{s}_j}^{\top}\mathbf{t}}{\tau_t}
    \right),
    \qquad
    p^{s}_j=\operatorname{softmax}
    \left(
        \frac{\mathbf{s}_j^{\top}\mathbf{t}}{\tau_s}
    \right),
    \label{eq:rel-dist}
\end{equation}
where $\tau_t$ and $\tau_s$ are the teacher and student temperatures, respectively.
We then transfer the relational structure by defining
\begin{equation}
d\left(\mathbf{s}_j^{},\hat{\mathbf{s}}_j\right)=\mathrm{KL}\left(p_{j}^{t},\Vert,p_{j}^{s}\right).
\label{eq:rel-loss}
\end{equation}

Thus, the relational objective transfers the teacher's relative semantic relationships with the text features rather than enforcing pointwise equality between $\mathbf{s}_j^{(k)}$ and $\hat{\mathbf{s}}_j^{(k)}$. The student is therefore not required to reproduce the absolute coordinates of the teacher representation. 
In practice, we compute the relational objective using batchwise image-text 
We form the matrices over the global batch to provide each visual representation with a larger set of text features for relational comparison. 

\vspace{-0.3cm}
\section{Experiments}
\label{sec:exp}

\begin{table}[t]
\caption{
\textbf{Main results across five task families.} All models use a ViT-B
backbone; teacher rows (ViT-L) are shown as per-task ceilings. Metric directions
are marked in the header, best values are in bold and second-best underlined, and
``-'' indicates the model lacks that capability. $^{\ddagger}$~Segmentation for
RADIOv2.5 is distilled from SAM, and for SigLino and EUPE from DINOv3.
$^{\S}$~DUNE result obtained by fine-tuning on non-public data.
}
\vspace{-0.5cm}
\label{tab:main}
\begin{center}
\resizebox{\textwidth}{!}{%
\begin{tabular}{llcccccccc}
\toprule
& & \textbf{Image Understanding} & \multicolumn{2}{c}{\textbf{2D Dense Prediction}} & \multicolumn{2}{c}{\textbf{3D Human Perception}} & \textbf{3D Vision} & \multicolumn{2}{c}{\textbf{Vision-Language}} \\
\cmidrule(lr){3-3}\cmidrule(lr){4-5}\cmidrule(lr){6-7}\cmidrule(lr){8-8}\cmidrule(lr){9-10}
\textbf{Model} & \textbf{Arch.} & \textbf{IN-1k cls} $\uparrow$ & \textbf{Depth} $\downarrow$ & \textbf{mIoU} $\uparrow$ & \textbf{F1} $\uparrow$ & \textbf{PA-PVE} $\downarrow$ & \textbf{AUC} $\uparrow$ & \textbf{COCO (T$\rightarrow$I/I$\rightarrow$T)} $\uparrow$ & \textbf{Flickr (T$\rightarrow$I/I$\rightarrow$T)} $\uparrow$ \\
\midrule
\multicolumn{10}{l}{\emph{Teachers (ceilings)}} \\
DINOv2      & ViT-L & --            & 0.384 & 47.70 & --   & --   & --   & --   & -- \\
Multi-HMR   & ViT-L & --            & --    & --    & 95.0 & 36.9 & --   & --   & -- \\
MASt3R      & ViT-L & --            & --    & --    & --   & --   & 91.2 & --   & -- \\
SigLIP2     & ViT-L & 83.8          & --    & --    & --   & --   & --   & 55.7 / 72.1 & 83.2 / 95.1 \\
\midrule
\multicolumn{10}{l}{\emph{Agglomerative baselines}} \\
RADIOv2.5   & ViT-B  & 72.6          & 0.718 & 50.0$^{\S}$  & --   & -- & -- & 48.9 / 63.2 & 76.2 / 89.7 \\
DUNE        & ViT-B  & --           & \textbf{0.358} & 45.6  & \textbf{94}   & \textbf{56.0} & \textbf{94.7}$^{\ddagger}$ & -- & -- \\
SigLino     & ViT-B  & {80.5}        & 0.750$^{\S}$    & \textbf{52.9}$^{\S}$  & --   & --   & --   & \textbf{55.6} / \textbf{72.9} & \textbf{81.9} / \textbf{94.2} \\
EUPE        & ViT-B  & 79.7          & \underline{0.391}$^{\S}$ & \underline{52.4}$^{\S}$ & -- & -- & -- & -- & -- \\
\midrule
\multicolumn{10}{l}{\emph{GRAFT models}} \\
$\student{1}$ DINOv2              & ViT-B  & --   & 0.420 & 45.5 & --   & --   & --   & --   & -- \\
$\student{2}=\student{1}+$ Multi-HMR     & ViT-B  & --   & 0.429 & 45.5 & 89   & 70.8 & --   & --   & -- \\
$\student{3}=\student{2}+$ MASt3R    & ViT-B  & --   & 0.433 & 45.7 & 89   & 75.6 & {84.2} & --   & -- \\
$\student{4}=\student{3}+$ SigLIP2        & ViT-B  & {81.2} & 0.441 & 46.0 & 89   & {78.1} & 81.8 & 51.9 / 69.0 & 76.2 / 88.4 \\
\textbf{GRAFT (ours)} & ViT-B & \textbf{81.1} & 0.437 & 46.0 & \underline{89} & \underline{75.3} & \underline{82.4} & \underline{52.3} / \underline{69.7} & \underline{76.8} / \underline{89.7} \\
\bottomrule
\end{tabular}%
}
\vspace{-0.8cm}
\end{center}
\end{table}
\vspace{-0.2cm}
\subsection{Datasets and metrics}
\label{sec:exp:setup}
\vspace{-0.2cm}

\noindent\textbf{Teachers models.} We distill four heterogeneous teachers, each a ViT-L encoder. 
DINOv2~\citep{oquab2024dinov2} is a self-supervised dense teacher whose features transfer to segmentation and depth. 
Multi-HMR~\citep{baradel2024multihmr} is for 3D human pose and mesh recovery. MASt3R~\citep{leroy2024mast3r} is for two-view 3D geometry and matching. SigLIP2~\citep{tschannen2025siglip2}, an image--text model that aligns visual and language representations.  
The default vision--language teacher is SigLIP2, with DFN5B~\citep{fang2024dfn} compared head to head in \S\ref{sec:exp:ablations}. 

\noindent\textbf{Datasets.} Distillation uses datasets drawn from a mixture that spans all four teacher domains. General 2D understanding and vision--language draw on DataComp-12M \citep{gadre2023datacomp}, ImageNet-21k \citep{ridnik2021imagenet21k}, Mapillary Street-level Sequences \citep{warburg2020msls}, and Google Landmarks Dataset v2 \citep{weyand2020gldv2}. 3D human perception draws on BEDLAM \citep{black2023bedlam}, AGORA \citep{patel2021agora}, and CUFFS \citep{baradel2024multihmr}. 3D geometry draws on ARKitScenesV2 \citep{baruch2021arkitscenes}, BlendedMVS \citep{yao2020blendedmvs}, CO3Dv2 \citep{reizenstein2021co3d}, DL3DV \citep{ling2024dl3dv}, Habitat512 \citep{savva2019habitat}, MegaDepthDense \citep{li2018megadepth}, Niantic \citep{arnold2022mapfree}, TartanAir \citep{wang2020tartanair}, Unreal4K \citep{tosi2021unreal4k}, VirtualKITTI \citep{cabon2020vkitti2}, and WildRGB \citep{xia2024wildrgbd}. 

\noindent\textbf{Implementation details.} The student model is a ViT-B/14 encoder at $448$px.
We evaluate every task family with its standard protocol, spanning the five domains GRAFT unifies. Following previous method~\citep{sariyildiz2025dune}, we drop the projectors and finetune a head for each downstream task. Image understanding is measured by ImageNet-1k classification top-1 accuracy. Dense prediction is measured by ADE20K linear-probe segmentation (mIoU) and monocular depth (RMSE). 3D human pose is measured on BEDLAM by F1 and PA-PVE. 3D geometry uses the
MASt3R-based suite, covering map-free relocalization, multi-view depth, and multi-view pose. For the 3D-geometry evaluations we substitute the
GRAFT encoder for the MASt3R encoder, re-initialize the encoder--decoder linear layer, and fine-tune the decoder and prediction heads. Vision--language is measured by COCO-5K and Flickr-30K retrieval in both directions, at R@1. We substitute the GRAFT encoder for the visual encoder and keep the same text encoder.

\begin{table}[t]
\caption{
\textbf{Ablation studies for vision language distillation.}
We study the distillation strategy, vision language projector architecture, and choice of vision language teacher. All experiments start from a student distilled from DINOv2 and further distill an incoming vision language teacher.
}
\label{tab:ablations}
\centering
\resizebox{\textwidth}{!}{
\begin{tabular}{lccc|lccc}
\toprule
& \textbf{Acquisition} & \multicolumn{2}{c|}{\textbf{Retention}} &
& \textbf{Acquisition} & \multicolumn{2}{c}{\textbf{Retention}} \\
\cmidrule(lr){2-2}\cmidrule(lr){3-4}\cmidrule(lr){6-6}\cmidrule(lr){7-8}
\textbf{Variant} & \textbf{IN-1k cls} $\uparrow$ & \textbf{mIoU} $\uparrow$ & \textbf{Depth RMSE} $\downarrow$ & 
\textbf{Variant} & \textbf{IN-1k cls} $\uparrow$ & \textbf{mIoU} $\uparrow$ & \textbf{Depth RMSE} $\downarrow$ \\
\midrule
\multicolumn{4}{l|}{\emph{Distillation strategy}} & 
\multicolumn{4}{l}{\emph{Vision--language projector}} \\
Direct distillation          & \textbf{82.6} & 35.4 & 0.584 & Frozen encoder + Transformer & 75.5 & 45.5 & \textbf{0.425} \\
LoRA finetuning        & 76.2 & 45.7 & \textbf{0.425} & Transformer, 1 layer & 80.4 & 46.0 & 0.448 \\
LPFT  & 78.0 & 40.5 & 0.442 & Transformer, 3 layers & 75.9 & 45.4 & 0.451 \\
EWC                    & 77.1 & 40.2 & 0.508 & \textbf{MLP + residual (ours)} & \textbf{81.2} & \textbf{46.0} & {0.450} \\
\cmidrule{5-8}
GRAFT w/o TSRT \& Relational Loss         & 77.4 & 40.6 & 0.489 & \multicolumn{4}{l}{\emph{Vision--language teacher}} \\
GRAFT w/o Relational Loss                & 80.1 & 43.6 & 0.455 & DFN5B (softmax InfoNCE)     & 80.0 & 45.7 & {0.454} \\
\textbf{GRAFT (ours)}       & {81.2} & \textbf{46.0} & {0.450} & 
\textbf{SigLIP2 (pairwise sigmoid, ours)} & \textbf{81.2} & \textbf{46.0} & \textbf{0.450} \\
\bottomrule
\end{tabular}
}
\vspace{-0.2cm}
\end{table}

\vspace{-0.15cm}
\subsection{Main results}
\label{sec:exp:main}
\vspace{-0.15cm}
\noindent\textbf{Performance across different domains.}
GRAFT is the only model in Table~\ref{tab:main} that reports a result in every column: a single ViT-B backbone spanning image understanding, 2D dense prediction, 3D human perception, 3D vision, and vision--language. Each baseline covers only the families its teacher set targets and is blank elsewhere. This breadth does not cost strength. On image understanding GRAFT attains the strongest classification of any agglomerative baseline. On 2D dense prediction it stays competitive with the segmentation- and depth-specialized baselines. On 3D human perception it trails only DUNE, the sole baseline that targets pose. On 3D vision it acquires geometric capability that most baselines lack outright. On vision--language it is second only to the vision--language-specialized baseline and ahead of every other agglomerative model that attempts retrieval. GRAFT therefore leads on image understanding and matches or approaches the strongest specialized baseline on every remaining family, while being the only model to cover all of them.

Table~\ref{tab:main} places GRAFT beside baselines trained under materially different conditions, so it should not be read as a controlled ranking. Two differences dominate, detailed in Appendix~\ref{app:conditions}. First, the teacher sets differ in number and scale: several baselines distil from far larger teachers, up to $\sim$1.1B parameters for RADIOv2.5 and $1.9$B for EUPE, whereas every GRAFT teacher is a ViT-L of $\sim$304M. Second, the training corpora differ in size and composition; several baseline corpora are undisclosed, and SigLino reports image tokens rather than images. Because a stronger teacher or a larger, better-matched corpus directly lifts the metrics Table~\ref{tab:main} tabulates, a baseline's lead on a given family reflects its setup as much as its method. GRAFT also carries the leanest budget of the five, never training its trunk from scratch: it warm-starts from the DUNE base and re-pays no base-encoder cost per graft. The controlled comparisons that hold teacher, data, and schedule fixed are reported in \S\ref{sec:exp:ablations}; Table~\ref{tab:main} instead makes the coverage point, that one ViT-B under a single fixed teacher and data budget answers in every column.

\noindent\textbf{Continually growing the model from a single teacher.}
The main-table GRAFT warm-starts from DUNE and performs a single vision--language graft (\S\ref{sec:exp:setup}). 
This ablation removes that convenience and asks the harder question the setting poses: can GRAFT \emph{build} the agglomerative model itself, one teacher at a time, with no pre-assembled base? 
Starting from a student distilled from a single dense teacher (DINOv2), we graft Multi-HMR, then MASt3R, then SigLIP2, and evaluate the student $\student{k}$ after every graft in Table~\ref{tab:main}. The table reads two ways. Across a row, each freshly grafted capability switches on at a level comparable to the warm-started model. Down a column, a capability introduced at one step persists through every later graft rather than collapsing: dense segmentation is retained in full and even edges up, depth drifts only modestly, pose detection holds while vertex error drifts within a bounded range, and 3D-vision accuracy gives back only a little after the vision--language graft. The four-capability model built entirely by continual grafting reaches performance comparable to the DUNE-warm-started GRAFT, trailing it only slightly on 3D pose and geometry. 

\vspace{-0.2cm}
\subsection{Ablations}
\label{sec:exp:ablations}
\vspace{-0.2cm}

\noindent\textbf{Distillation strategy.}
In Table~\ref{tab:ablations}, we compare GRAFT with alternative strategies for incorporating a new vision-language teacher into an already distilled student from DINOv2. Direct distillation directly optimizes the student toward the new teacher and achieves strong acquisition of the new capability, but substantially degrades the previously distilled dense and geometric capabilities, illustrating the stability plasticity conflict in sequential teacher integration. LoRA~\citep{hu2021lora} finetuning instead introduces a portable task specific adaptation. We evaluate retention with the LoRA parameters removed, which recovers the original student behavior, but the resulting acquisition of the new capability remains limited. We further consider linear probe then finetune (LPFT)~\citep{kumar2022fine} and EWC based~\citep{kirkpatrick2017ewc} continual learning, both of which improve the balance between acquisition and retention but still exhibit noticeable interference with previously learned capabilities. Starting from the basic GRAFT pipeline without TSRT or the relational objective, we observe a clear stepwise improvement as each proposed component is introduced. 
These results support the complementary roles of task specific readouts and relational distillation in integrating heterogeneous teachers without disrupting previously acquired representations.

\begin{figure}[t]
    \centering
    \includegraphics[width=1\linewidth]{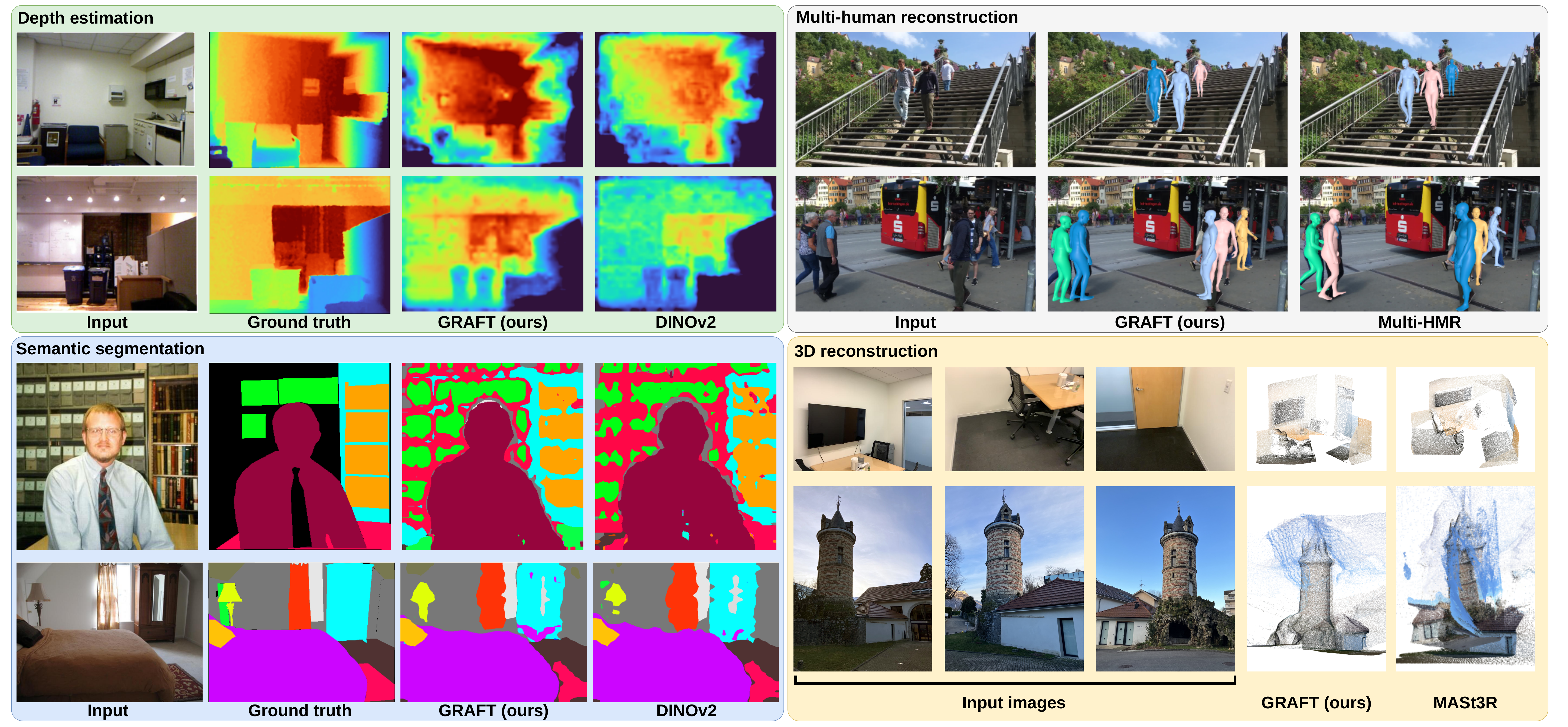}
    \vspace{-0.8cm}
    \caption{\textbf{Qualitative comparison} on depth estimation, semantic segmentation, multi-human reconstruction, and 3D reconstruction between GRAFT and corresponding teachers.}
    \label{fig:qualitative}
    \vspace{-0.6cm}
\end{figure}

\noindent\textbf{Choice of projector.}
We study the architecture of the projector. One option is to attach a Transformer projector and optimize only the newly introduced head while keeping the distilled encoder frozen. This largely avoids modifying the previously distilled backbone and therefore preserves part of its original capabilities, but provides insufficient flexibility for acquiring the new vision-language representation. 
Increasing the Transformer depth does not consistently improve performance, suggesting that a more expressive projector alone does not resolve the representation mismatch between the student and the vision language teacher. Our final design instead uses a lightweight MLP projector with a residual connection. Despite its substantially simpler architecture, it achieves acquisition and retention comparable to or better than the Transformer based alternatives across the evaluated tasks. This indicates that the benefits of GRAFT primarily arise from the proposed representation and optimization design rather than from increasing projector capacity.

\noindent\textbf{Vision language teacher.}
We examine whether the effectiveness of GRAFT depends strongly on the choice of vision-language teacher. In addition to SigLIP2, we replace the supervising model with DFN5B, which has a comparable model scale and capability.
GRAFT exhibits similar acquisition and retention behavior with both teachers.
This result suggests that the proposed grafting mechanism is robust to the specific choice of vision language teacher and can transfer vision-language knowledge without relying on properties unique to SigLIP2.

\vspace{-0.2cm}
\subsection{Qualitative results}
\label{sec:exp:qualitative}
\vspace{-0.2cm}

We provide qualitative comparisons between GRAFT and its corresponding teacher models across four downstream tasks: depth estimation, semantic segmentation, multi-human reconstruction, and 3D reconstruction. As shown in Figure~\ref{fig:qualitative}, our method achieves competitive or superior performance.
\vspace{-0.5cm}
\section{Conclusion}
\label{sec:conclusion}
\vspace{-0.25cm}

We introduced {GRAFT}, a continual multi-teacher distillation framework that enables agglomerative vision foundation models to grow beyond a fixed teacher set. Instead of revisiting all previously integrated teachers, GRAFT uses the previous student as a frozen retention teacher and distills jointly from it and the incoming foundation model, allowing new capabilities to be acquired while preserving prior ones at the cost of a single incremental distillation stage. To accommodate heterogeneous representation geometries, GRAFT further introduces Teacher Specific Readout Tokens and a Geometry Agnostic Relational Loss that transfers vision--language knowledge through image--text similarity structure rather than direct feature matching. Using this formulation, a single ViT-B backbone progressively integrates image understanding, 2D dense prediction, 3D human pose estimation, 3D vision, and vision--language capabilities.

\bibliography{iclr2027_conference}
\bibliographystyle{iclr2027_conference}

\clearpage
\appendix

\section{Comparison conditions}
\label{app:conditions}

The main comparison in Table~\ref{tab:main} is not strictly like-for-like: the methods differ in the teachers they distill and in the corpora they train on, both of which directly move the reported metrics.
Table~\ref{tab:ledger} collects these conditions.

\begin{table}[h]
\caption{\textbf{Teacher sets and training data differ across methods}, so the head-to-head in Table~\ref{tab:main} is not like-for-like. Teacher scale is per model and counts are the number of distillation teachers. Entries are stated by each source where available and marked unavail when a source does not disclose them. SigLino reports image tokens rather than images, and EUPE additionally trains a from-scratch 1.9B proxy before distillation. GRAFT's teachers are DUNE's three, inherited through the warm start, plus the grafted vision--language teacher; its data subtotal counts DataComp-12M and ImageNet-19k and excludes Mapillary and GLDv2, whose sizes are undisclosed.}
\label{tab:ledger}
\begin{center}
\resizebox{\textwidth}{!}{
\begin{tabular}{lll}
\toprule
\textbf{Method} & \textbf{Teachers (count $\times$ scale)} & \textbf{Training data} \\
\midrule
RADIOv2.5 & $4\times$ up to $\sim$1.1B (DINOv2-g, SAM-H, DFN-CLIP-H) & unavail \\
DUNE      & $3\times$ ViT-L/14 ($\sim$304M)                          & $\sim$20.7M images \\
SigLino   & $2\times$ ViT-L                                          & 230B image tokens \\
EUPE      & PEcore-G 1.9B, PElang-G 1.7B, DINOv3-H$+$ 840M ($+$1.9B proxy) & unavail \\
\midrule
\textbf{GRAFT (ours)} & $3\times$ ViT-L/14 ($\sim$304M) $+$ SigLIP2 (ViT-L) & $\sim$21.7M images \\
\bottomrule
\end{tabular}
}
\end{center}
\end{table}

\section{Extended results}
\label{app:tables}

\paragraph{3D geometry: map-free relocalization.}
Table~\ref{tab:mapfree} examines GRAFT's 3D-geometry capability in detail on the map-free relocalization leaderboard, with the GRAFT encoder substituted into MASt3R and only the decoder fine-tuned. GRAFT ranks second only to the MASt3R teacher it distills from, ahead of the dedicated matching and two-view baselines on virtual-correspondence AUC, reprojection error, and relative-pose AUC. The residual gap to the teacher concentrates in fine-grained pose precision and reflects the setting: GRAFT compresses a ViT-L geometry specialist into a ViT-B trunk shared with four other capabilities and adapts no backbone weights at evaluation. That a shared, unadapted trunk still recovers most of a dedicated matcher's relocalization accuracy shows the geometry graft leaves genuinely usable spatial structure in the trunk.

\begin{table}[h]
\caption{\textbf{Map-free visual relocalization}, official leaderboard (test
split). The GRAFT encoder is substituted for the MASt3R encoder and the MASt3R
decoder is fine-tuned, with the backbone frozen. AUC and precision (Prec.) are
percentages, and metric directions are marked in the header. Baseline rows are
quoted from \citet{sariyildiz2025dune}, Table~4. 
}
\label{tab:mapfree}
\begin{center}
\resizebox{\textwidth}{!}{%
\begin{tabular}{llccccccccc}
\toprule
& & \multicolumn{2}{c}{\textbf{VCRE} $<45$\,\textbf{px}} & \multicolumn{2}{c}{\textbf{VCRE} $<90$\,\textbf{px}} & \textbf{Median reproj.} & \multicolumn{4}{c}{\textbf{Relative pose error}} \\
\cmidrule(lr){3-4}\cmidrule(lr){5-6}\cmidrule(lr){7-7}\cmidrule(lr){8-11}
\textbf{Method} & \textbf{Encoder} & \textbf{AUC} $\uparrow$ & \textbf{Prec.} $\uparrow$ & \textbf{AUC} $\uparrow$ & \textbf{Prec.} $\uparrow$ & \textbf{error (px)} $\downarrow$ & \textbf{Trans.\ (m)} $\downarrow$ & \textbf{Rot.\ ($^\circ$)} $\downarrow$ & \textbf{Prec.} $\uparrow$ & \textbf{AUC} $\uparrow$ \\
\midrule
LoFTR \citep{sun2021loftr}              & CNN   & 39.7 & 18.2 & 61.8 & 33.5 & 166.8 & 2.31 & 39.4 & 26.9 & 9.8 \\
DUSt3R \citep{dust3r2024}               & ViT-L & 45.9 & 28.7 & 69.8 & 50.4 & 115.8 & 0.99 & 7.1  & 39.4 & 21.4 \\
MicKey \citep{barrosolaguna2024mickey}  & ViT-L & 57.2 & 31.2 & 74.8 & 49.3 & 129.5 & 1.59 & 26.0 & 28.3 & 12.0 \\
MASt3R \citep{leroy2024mast3r}          & ViT-L & \textbf{81.7} & \textbf{63.0} & \textbf{93.3} & \textbf{79.3} & \textbf{48.8}  & \textbf{0.37} & \textbf{2.2} & \textbf{74.0} & \textbf{54.7} \\
GRAFT ($\student{4}$, ours)         & ViT-B & \underline{60.9} & \underline{48.0} & \underline{82.2} & \underline{69.1} & \underline{73.1} & \underline{0.64} & \underline{7.1} & \underline{27.9} & \underline{38.8} \\
\bottomrule
\end{tabular}%
}
\end{center}
\end{table}

\end{document}